\documentclass{article}

\usepackage{PRIMEarxiv}

\usepackage[utf8]{inputenc} 
\usepackage[T1]{fontenc}    
\usepackage{hyperref}       
\usepackage{url}            
\usepackage{booktabs}       
\usepackage{amsfonts}       
\usepackage{nicefrac}       
\usepackage{microtype}      
\usepackage{lipsum}
\usepackage{fancyhdr}       
\usepackage{graphicx}       
\graphicspath{{media/}}     

\usepackage[font={small}]{caption}

\title{Using evolutionary computation to optimize task performance of unclocked, recurrent Boolean circuits in FPGAs
}

\author{
  Raphael Norman-Tenazas, David Kleinberg, Erik C. Johnson, Matthew J. Roos \\
  Johns Hopkins University Applied Physics Laboratory \\
  Laurel, MD\\
  \texttt{matt.roos@jhuapl.edu} \\
   \And
  Daniel P. Lathrop \\
  Recurrent Computing Inc. \\
  Reston, VA \\
}

\begin{document}
\maketitle

\begin{abstract}
It has been shown that unclocked, recurrent networks of Boolean gates in FPGAs can be used for low-SWaP reservoir computing. In such systems, topology and node functionality of the network are randomly initialized. To create a network that solves a task, weights are applied to output nodes and learning is achieved by adjusting those weights with conventional machine learning methods. However, performance is often limited compared to networks where all parameters are learned. Herein, we explore an alternative learning approach for unclocked, recurrent networks in FPGAs. We use evolutionary computation to evolve the Boolean functions of network nodes. In one type of implementation the output nodes are used directly to perform a task and all learning is via evolution of the network’s node functions. In a second type of implementation a back-end classifier is used as in traditional reservoir computing. In that case, both evolution of node functions and adjustment of output node weights contribute to learning. We demonstrate the practicality of node function evolution, obtaining an accuracy improvement of $\sim$30\% on an image classification task while processing at a rate of over three million samples per second. We additionally demonstrate evolvability of network memory and dynamic output signals.
\end{abstract}


\section{Introduction}
Under traditional reservoir computing (RC) a spatial or spatiotemporal signal is passed into a “reservoir” of artificial neurons (nodes)\cite{b1, b2, b3}. This reservoir is a graph network of randomly connected nodes with random connection weightings and non-linear activation functions. Typically, the number of reservoir nodes is higher than the dimensionality of the input signal. Additionally, the recurrent topology of the reservoir generates temporal dynamics within the network, which, even for a static input signal, further increases overall dynamical spatiotemporal dimensionality. Some or all of the network nodes also serve as output nodes. As the network responds to an input signal, signals from the output nodes are captured across a fixed period of time. These output signals are then used as input to a “back-end” traditional machine learning model, such as a linear or logistic classifier. The back-end model is trained in a supervised manner, using labels of signals that are input to the network, paired with the captured output signals.

Recently it has been shown that RC may alternatively be implemented in field-programmable gate arrays (FPGAs) as unclocked, recurrently-connected Boolean logic gates\cite{b4, b5, b6, b7}. Although these gates are Boolean in nature, the recurrence, along with the non-instantaneous switching of these gates, can result in mixed analog-digital dynamic behavior, akin to traditional RCs with sigmoidal or hyperbolic tangent node activation functions. Such implementation allows for extremely low size, weight, and power (low-SWaP) processing compared to traditional embedded graphics processors.

In such FPGA implementations, look-up tables (LUTs) serve as network nodes, and the Boolean function implemented by a LUT determines the behavior of that node, in lieu of the weighted connections of artificial neurons in traditional RCs. LUTs are defined by a vector array of bits, and those bits are initialized randomly or using heuristics to target a desired network sensitivity\cite{b5, b7}. The network’s topology and node functionality are then fixed, and all future learning occurs via the weights applied to the outputs of the network, across space and time. Previous researchers have applied such FPGA-RC systems to both spatial tasks, such as image processing\cite{b5, b6}, and temporal tasks, such as RF signal classification\cite{b7}.



\begin{figure*}[ht!]
\centerline{\includegraphics[width=1.0\textwidth]{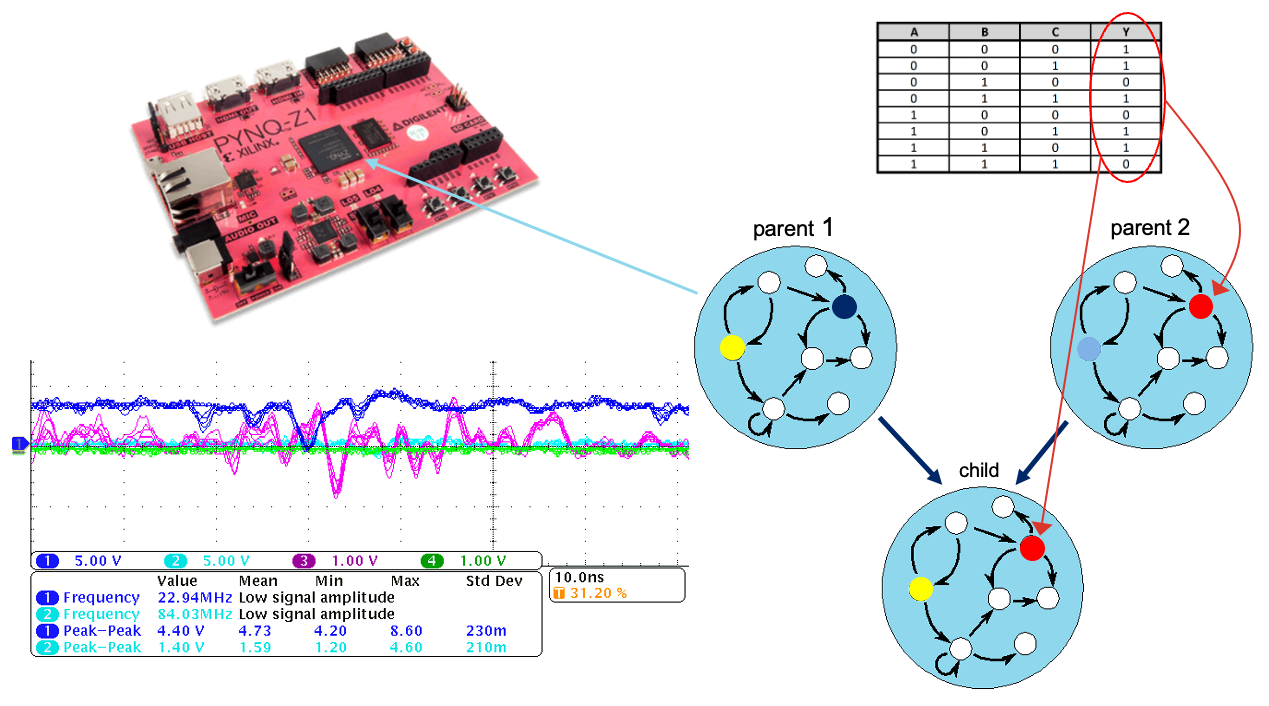}}
\caption{
We implemented unclocked, recurrent networks of Boolean gates (e.g., "parent 1") on a PYNQ-Z1 development board hosting a Xilinx Zynq FPGA. Network node functions are reconfigurable LUTs defined by a vector of bits (3-input, 1-output LUTs in this depiction, but 5-input, 1-output LUTs in our PYNQ implementations). Such networks exhibit nanosecond-scale analog-like dynamics (\textit{lower left}) that are not perfectly repeated for repeated digital inputs (\textit{overlayed traces}). The discrete nature of digital logic prevents the use of gradient descent in on-hardware network training. Instead, we use evolutionary computation, in which genes (LUTs) of parent networks are combined and mutated to produce child networks. Only the best performing child networks are then used as parents when creating the subsequent generation of networks.
}
\label{fig:overview}
\end{figure*}

Although it is convenient to train only the back-end model of an RC system, task performance may be sub-optimal if the random configuration of the reservoir is not well-suited for the task. In this work, we explored an alternative or augmented learning method in which the reservoir itself is adapted to the task. Namely, we used evolutionary computation to evolve the Boolean functions implemented by each node in the network. This method is motivated by the fact that gradient descent, the most popular modern approach to optimizing neural networks, cannot be implemented in our unclocked, recurrent networks, and thus a gradient-free method is necessary.

While training via evolutionary computation has been utilized before in conventional Von Neumann computer architectures\cite{b8}, the use of an unclocked, recurrent network of FPGA LUTs poses special challenges that we sought to overcome. Historically, changes to an FPGA network required the compilation of a new bitstream that is then loaded onto the FPGA---a process that can take hours to days for each network. Evolving a population of hundreds of networks for thousands of generations would thus require years of wall-clock time, if using a single FPGA. While one might attempt to simulate the quasi-analog system in software, this too would require extremely lengthy wall-clock processing time, and is also unlikely to adequately map to the nuances of the physical hardware. Our approach leverages reconfigurable LUTs (see Methods section) for fast network updates and on-hardware evaluation, enabling evolution within practical wall-clock times of hours to days.

\section{Methods}

Our objective was to demonstrate (1) the benefit of our training approach---namely, enhancing task performance compared to that of the more traditional reservoir computing (RC) paradigm when implemented in recurrent, unclocked Boolean circuits on FPGAs, and (2) the practicality of our training approach, that is, being able to do so within reasonable time periods (hours to days). We additionally aimed to demonstrate basic learning capabilities that may suggest applicability of our approach to a diverse set of spatial and temporal tasks.

An overview of previous approaches\cite{b4, b5, b6, b7}, along with our technical contribution, is depicted in Figure \ref{fig:overview}. In contrast to traditional RC, we alternatively or additionally allow for learning via adaptation of node functions. We use evolutionary computation to optimize node functions for a given task—testing a population of LUT configurations, selecting those that give best task performance, then cloning and mutating the best performing configurations to generate the next population. This is repeated for many generations.

Our hardware implementation used a PYNQ-Z1 board---a low-cost hobbyist development board containing a Xilinx Zynq FPGA. Our approach and developed software are equally applicable to industry-grade Xilinx FGPAs containing over $10^5$ LUTs (nodes) and capable of running extremely large networks potentially useful for real-world applications.

Importantly, our approach requires the use of reconfigurable LUTs. This hardware innovation is available on a subset of FPGAs including the Xilinx Zynq family of FPGAs, as the ``CFGLUT5'' element. It allows the bits that define a LUT’s function to be updated in less than a microsecond. Prior to the availability of reconfigurable LUTs, a new FPGA bitstream had to be compiled each time such bits were changed---taking hours to days.

We realized and tested two types of system implementations. In one type of implementation the network's output nodes are used directly to perform task---no back-end model is needed or used, and all learning is via evolution of the network’s node functions. In a second type of implementation a back-end model is used as in traditional reservoir computing. In this second type, both evolution of node functions and adjustment of output node weights contribute to learning. We demonstrate the practicality of evolution of node functions, and compare task performance of RCs with and without evolution.

\section{Experiments and Results}

Here we present results from three sets of experiments. In the first experiment, we demonstrate the benefit of our use of evolutionary computation to train a network, compared to the traditional reservoir computing approach. The second and third experiment demonstrate general learning capabilities, which are likely to be beneficial or be required for certain types of real-world tasks.

\begin{figure*}[ht!]
\centerline{\includegraphics[width=0.8\textwidth]{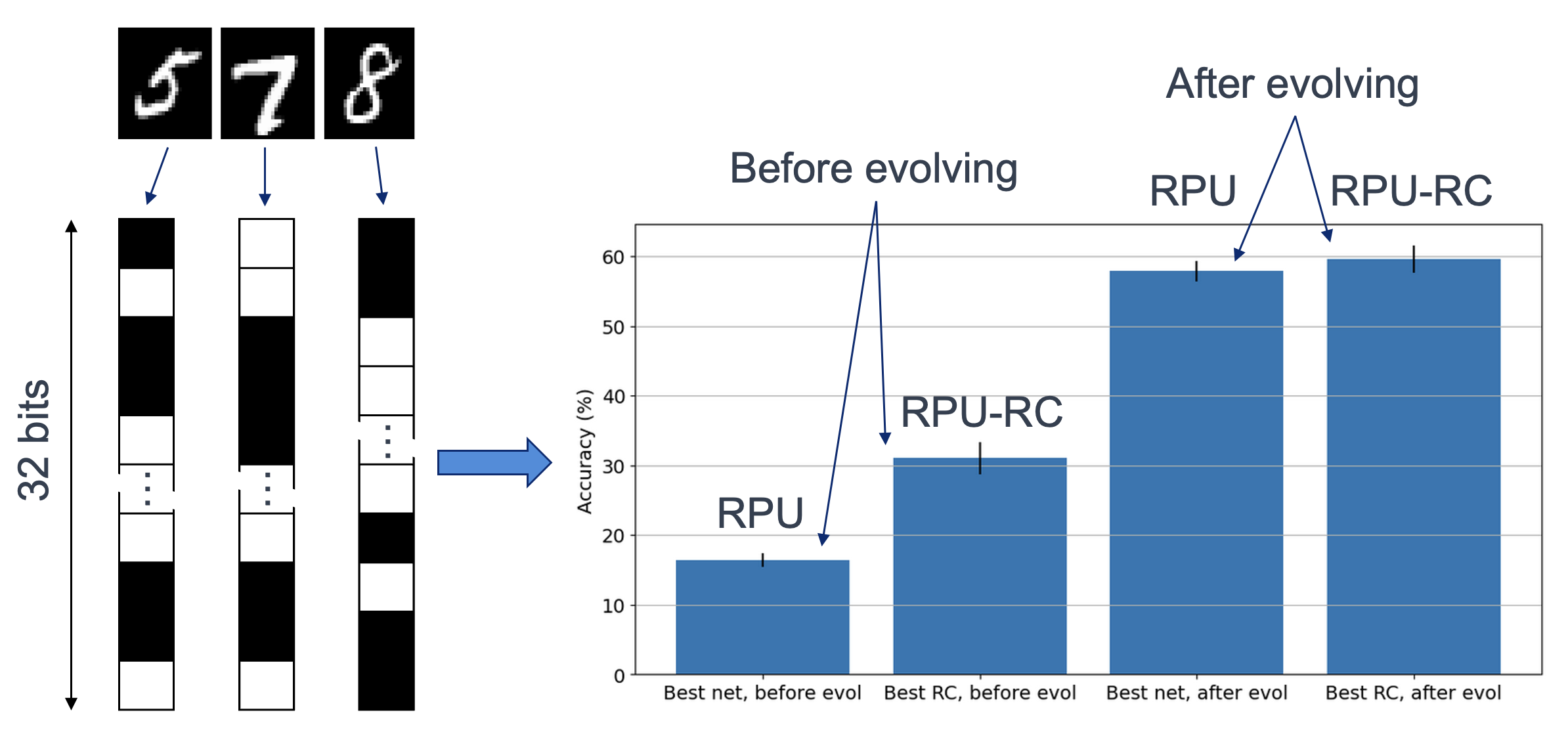}}
\caption{
\textit{Left}: Handwritten digits of 784 (28x28) pixels are converted to 32-bit representations by discarding the least informative bits. This results in some information loss, but is done to accommodate use of the low-cost, low-resource PYNQ development board. \textit{Right}: A network of 100 LUTs is used to process input digits (vectors) and make classification predictions. Before evolution, the traditional reservoir computing paradigm (“RPU-RC”), which uses a trained back-end classifier, performs better than a naïve network with no back-end classifier (“RPU”). After evolution, both RPU and RPU-RC perform much better than before evolution. After evolution, the use of the back-end classifier in the RPU-RC provides little benefit over the RPU. Nearly all processing is done internally by the dynamic interactions of the recurrent network, and the activation of one of ten output nodes directly indicates the classification prediction.
}
\label{fig:mnist}
\end{figure*}

\subsection{Image classification}
In this task we trained a network of 100 LUTs to classify 32-bit representations of handwritten digits. The digits are 784 (28x28) pixel images taken from the MNIST database. To accommodate the limited resources on the PYNQ board, we reduced these images to 32 bits through the following heuristic procedure.
\begin{enumerate}
  \item Pixel values were binarized by thresholding at 0.5.
  \item The 784 bits were partitioned into 49 sets of 16 bits (identical partitioning for all images).
  \item For each of the 49 sets, the 16-bit representations were used to train a neural network model with training set images (90 hidden nodes with ReLU activation, 10 output nodes with softmax activation, cross-entropy loss function, 15 training epochs). Accuracy was then computed on the test set images.
  \item Steps 2 and 3 were repeated 10 times, using different random partitioning each time.
  \item For each bit, we computed the mean of the 10 accuracy scores to which that bit contributed, and then rank-ordered those scores.
  \item The 32 bits with the highest accuracy scores from step~5 were chosen to represent the MNIST digits. The other 752 bits were discarded.
\end{enumerate}

Note that the above approach results in some information loss, such that even a top performing traditional neural network could not achieve greater than $\sim$87\% accuracy on the 32-bit representations.

The topology of the 100-LUT network was randomly created with the exception that it was assured that (1) each of the 5-input LUTs received five inputs (none were left unconnected), (2) 32 of the LUTs received one and only one of the 32 input bits, and (3) 10 LUTs provided output bits used for classification directly, or for input to a back-end logistic regression classifier. None of the output LUTs directly received an input bit.

Input 32-bit vectors were assembled into groups of 2000 and fed into a network sequentially at 3.125 MHz, one vector every 0.32 $\mu$s, while the 10 output LUTs were sampled with a 1-bit ADC at 100 MHz. This results in 32 output samples (across time) per input digit per output LUT.

Classification was done in two ways---with or without a back-end classifier. When no back-end classifier is used, nearly all processing is done internally by the dynamic interactions of the recurrent network, and the activation of one of ten output nodes directly indicates the classification prediction. We thus refer to the network as a Recurrent Processing Unit (RPU). In this approach, the stream of 32 output bits from each output LUT, which coincide with the duration of the presentation of an input digit, are averaged as floating-point numbers. The predicted digit (0-9) is that associated with the output LUT that gives the largest averaged output.

When a back-end classifier is used, we are implementing a reservoir computer (RC). Because the RPU as described above serves as the reservoir, we refer to this as an RPU-RC. Under this approach, the stream of 32 output bits from the 10 output LUTs give 320 bits that can be classified as one of ten digits by a back-end classifier trained using logistic regression. The back-end classifier is trained on the host computer using RPU outputs from all 60,000 MNIST training samples.

To train our networks using evolutionary computation, we used a fixed topology but initialized a population of 100 networks with randomly chosen LUT bits. In a given generation, each network was scored on a batch of 2000 digits. The 20 top performing networks were used as parents, and the top four were preserved unmutated for the next generation. For a given child, two parents are randomly chosen from the top performing network pool. Each of the child’s LUTs were randomly selected from one of the two parents (i.e., bits within a given LUT are from one and only one of the two parents). Then, mutations were induced by flipping individual LUT bits with a probability of 0.0033. We evolved our networks over 10,000 generations, which took several hours. Our post-processing is done on a host machine, resulting in a bottleneck between the PYNQ and host machine. In future work, this bottleneck could be removed by implementing post-processing on the development board itself.

Results are shown in Figure \ref{fig:mnist}. Scores are shown only for the highest performing network within the population of 100 networks. Prior to evolution, we see that the RPU-RC system performs much better than the RPU alone, thanks to the training of the RPU-RC’s back-end classifier. Nonetheless, RPU-RC performance is relatively poor, at only $\sim$31\% accuracy. After evolution, scores for both RPU and RPU-RC improve substantially, up to almost 60\%. Notably, after evolution, the use of the back-end classifier in the RPU-RC provides little benefit over the RPU. The back-end classifier can be dropped, saving on processing time and energy.

Although 60\% accuracy is well below the 87\% accuracy of a traditional neural network, we note that larger networks and more evolution generations are likely to close that gap. In limited testing with a network of 500 LUTs, we observed accuracies of 70\% or higher (data not shown). An FPGA capable of implementing networks of thousands of LUTs very well may achieve 87\% accuracy, while still being energy efficient and extremely fast.

\subsection{Dynamic output}
For some types of tasks, it is desirable for a system to generate temporally dynamic output, such as in a control system. In this “digital-to-frequency conversion” task, we demonstrate the capability to train a network to generate different dynamic signals for a given static input. The network’s task is to take in four static bits that represent a number from 0 to 15. Note that these are not the MNIST digits as used in the experiment above, but 4-bit integers in most signficant byte (MSb) format. The network’s single output bit should fluctuate at a rate that is correlated to the value of the input number. That is, an input of [0, 0, 0, 0] (decimal 0) should generate an output that fluctuates slowly or not at all while an input of [1, 1, 1, 1] (decimal 15) should generate an output that fluctuates more rapidly than for any other input.

We used a smaller network (24-LUTs) than in the image recognition task, but the same population size (100) and number of parents (20). Across 100 generations of evolution, each network was stimulated with digits from 0 to 15 (four-bit representations), presented in random order. Outputs were sampled with the 1-bit ADC and the number of transitions (0 to 1 or 1 to 0) were counted over the period during which each of the inputs digits was presented. The Pearson correlation coefficient between the output fluctuation rates (“output frequency”) and the input MSb integer values were used as fitness scores.

\begin{figure}[htbp]
\centerline{\includegraphics[width=0.5\textwidth]{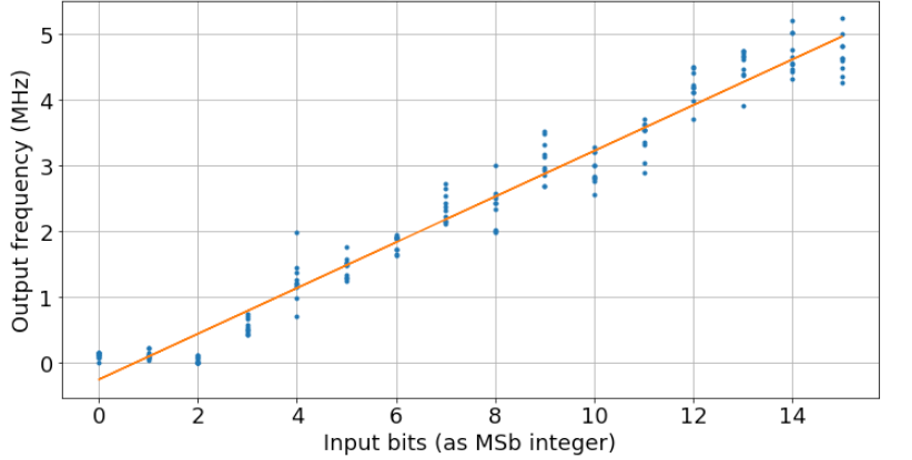}}
\caption{
Performance of a network evolved to perform “digital-to-frequency” conversion. The 24-LUT network’s task is to take in a 4-bit integer (as four separate binary inputs, and representable as the decimal numbers 0 to 15) and generate a single output that fluctuates at a rate correlated with the value of the input decimal number. For this trained network, the correlation coefficient between the input values and the output fluctuation rates (“output frequency”) was 0.98. This experiment demonstrates the ability to evolve networks that have targeted dynamic outputs—a capability necessary for certain types of tasks such as operation of a control system.
}
\label{fig:dynamic}
\end{figure}

As seen in Figure \ref{fig:dynamic}, a network of only 24 LUTs can achieve high correlation. Maximum performance was reached quickly, in less than 50 generations. Although repeated trials exhibit variation about a mean for a given input, we speculate that a larger network with more LUTs and more training generations will exhibit lower variation.

\subsection{Temporal memory}
For the final task, we directly demonstrate the capability to evolve a network that has temporal memory, and requires the use of that memory to execute a task. This capability is necessary for many time-domain tasks such as speech processing or RF signal classification. We evolved the network perform an “N-back” task, which is often used in psychology and cognitive neuroscience. In this one-bit version, the network is feed a sequence of single bits and the network must output those same bits, in the same order, with a delay of N time units.

\begin{figure}[htbp]
\centerline{\includegraphics[width=0.5\textwidth]{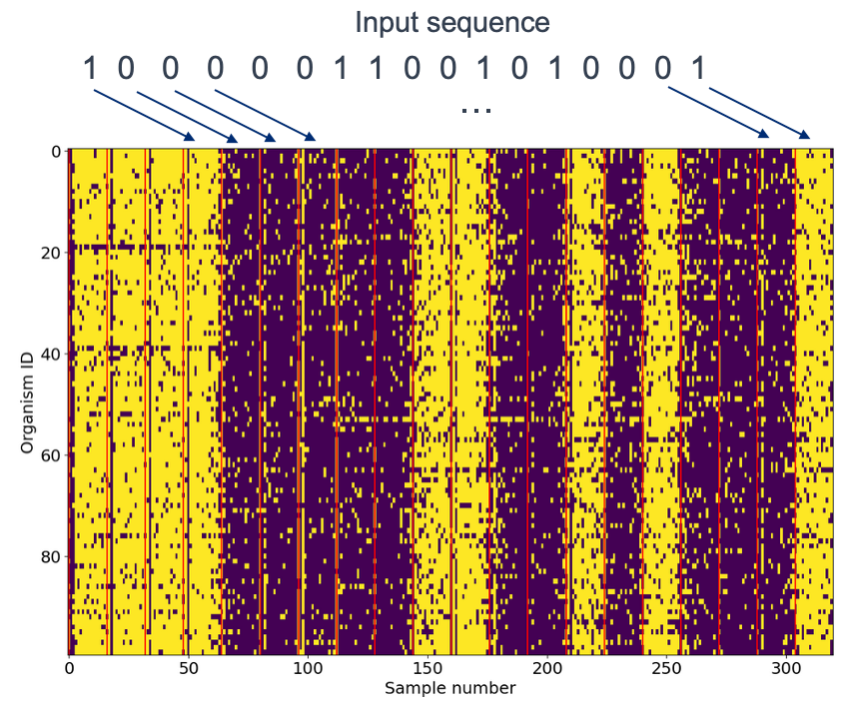}}
\caption{
Behavior of the final generation of networks evolved to perform an N-back task, with N=3. Each row is the output bit of one of 100 networks, while performing the task. Time (sample number) is from left to right. Input bits were each presented for 0.16 $\mu$s (16 samples of the output node), and transitions are indicated by red vertical lines. The input bit presented between each pair of red lines is the number printed above that column. Output bits are colored yellow (1) and blue (0). As desired, outputs are equal to inputs (when averaged over the duration of an input bit) except with a delay of 3 input digits. Nearly all members of the population perform well, although one poorly-performing child can be observed near row (network/organism) 55. Variation across rows is due to differences in LUT tables and to electronic noise owing to the analog nature of the circuit. This experiment demonstrates the ability to evolve networks that have intrinsic memory---a capability necessary for certain types of tasks such as speech recognition and RF signal classification.}
\label{fig:memory}
\end{figure}

Again using 20 parents and keeping the best four networks for the subsequent generation, a network of 100 LUTs evolved to achieve 100\% accuracy in under 250 generations with N=3 (Figure \ref{fig:memory}). In additional testing, networks achieved nearly perfect performance with N as high as 7 or 8 (data not shown). It is worth highlighting that the mechanism underlying the memory is the intrinsic capacitance and propagation time delays of the network’s underlying hardware. No RAM or similar storage media are used.

\section{Conclusion}
These results demonstrate the increasingly attractive possibility of using low-SWaP networks of unclocked, recurrently-connected Boolean logic gates for targeted machine learning tasks. By taking advantage of the speed of reconfigurable LUTs, evolutionary computation can be used to increase task performance beyond that of traditional reservoir computing paradigms. Although performance boost was most clearly demonstrated on a static classification task, additional experiments demonstrated capabilities useful or necessary for other types of tasks, such as time-domain signal processing (requiring memory) and control of dynamic systems (requiring dynamic output). Notably, processing is extremely fast, and the necessary hardware is relatively inexpensive and widely available. In future studies we aim to increase the scale of the networks and test the approach on a variety of more challenging tasks including closed-loop control and classification of time-domain signals. We also aim to compare our implementations with those in traditional hardware, via direct measurements of size, weight, and power.

\end{document}